\documentclass[sigconf,review=false, prologue,table,xcdraw]{acmart}
\usepackage{algorithm}
\usepackage{algorithmic}
\usepackage{subcaption}

\usepackage{graphicx}
\usepackage{multirow}
\usepackage{xcolor}
\usepackage{easyReview}

\usepackage{tablefootnote}
\usepackage{threeparttable}
\usepackage{lipsum}
\newtheorem{definition}{Definition}
\AtBeginDocument{%
  \providecommand\BibTeX{{%
    \normalfont B\kern-0.5em{\scshape i\kern-0.25em b}\kern-0.8em\TeX}}}

\setcopyright{acmcopyright}
\copyrightyear{2018}
\acmYear{2018}
\acmDOI{XXXXXXX.XXXXXXX}

\acmConference[Conference acronym 'XX]{Make sure to enter the correct
  conference title from your rights confirmation emai}{June 03--05,
  2018}{Woodstock, NY}
\acmPrice{15.00}
\acmISBN{978-1-4503-XXXX-X/18/06}

\usepackage{comment}
\newtheorem{remark}{Remark}[section]
\begin{document}

\title{GraphViz2Vec: A Structure-aware Feature Generation Model to Improve Classification in GNNs}

\author{Shraban Kumar Chatterjee and Suman Kundu}
\authornote{Both authors contributed equally to this research.}
\email{{chatterjee.2, suman}@iitj.ac.in}


\renewcommand{\shortauthors}{}
\begin{abstract}
GNNs are widely used to solve various tasks including node classification and link prediction. Most of the GNN architectures assume the initial embedding to be random or generated from popular distributions. These initial embeddings require multiple layers of transformation to converge into a meaningful latent representation. While number of layers allow accumulation of larger neighbourhood of a node it also introduce the problem of over-smoothing. In addition, GNNs are inept at representing structural information. For example, the output embedding of a node does not capture its triangles participation. 
In this paper, we presented a novel feature extraction methodology GraphViz2Vec that can capture the structural information of a node's local neighbourhood to create meaningful initial embeddings for a GNN model. These initial embeddings helps existing models achieve state-of-the-art results in various classification tasks. Further, these initial embeddings help the model to produce desired results with only two layers which in turn reduce the problem of over-smoothing. The initial encoding of a node is obtained from an image classification model trained on multiple energy diagrams of its local neighbourhood. These energy diagrams are generated with the induced sub-graph of the nodes traversed by multiple random walks. The generated encodings increase the performance of existing models on classification tasks (with a mean increase of $4.65\%$ and $2.58\%$ for the node and link classification tasks, respectively), with some models achieving state-of-the-art results.
\end{abstract}



\maketitle
\section{Introduction}
Graph Neural Networks (GNN) have become very popular due to their application in different domains, from medical \cite{9933896} to mathematics \cite{feng-etal-2021-graphmr}. Two of the most popular tasks that GNN effectively solves are node classification and link prediction. In general, node classification require initial embedding that represents node features extracted from the domain knowledge. For example, consider a graph with tweets as the node with mentions relationship between the nodes, the initial embeddings may be the encoding of the text using Word2Vec \cite{Mikolov2013EfficientEO}. In many cases, specifically for certain applications, random initial embedding is used. For example, in link classification, initial node embeddings can be randomly initialized from popular distributions like normal distribution. However, the graph structure is not encoded into the feature set and it is up to the GNN to infuse the final embeddings with the task-devout structural information. GNNs collate the neighbour's information into a node using message passing and aggregation blocks. It is shown that GNNs cannot retain simple structural properties like the triangle participation of a node \cite{chamberlain2023graph}. Also, GNNs can assign the same encodings to two nodes with similar neighbourhood structures, ignoring the distance between the two nodes \cite{Srinivasan2020On}. These proves to be a challenge, especially in the link prediction task where two nodes can have similar neighbourhoods but have different probabilities of creating new links with other nodes. Many methods tried to solve these by adding precomputed neighbourhood information into the initial embeddings of a node, e.g., its triangle participation \cite{10.5555/3327345.3327423}, positional information \cite{NEURIPS2019_9d63484a}, etc., with limited success \cite{10.1609/aaai.v33i01.33014602}. 
Additionally, GNNs also suffer from the over-smoothing problem \cite{Oono2020Graph}. Decreasing the number of layers in the GNN \cite{you2020l2} may reduce the over-smoothing at the expense of model performance. 

In this paper, we propose a novel non-message passing batched technique to generate implicit structure-aware input feature representation for nodes in a graph. These implicit features will retain all structural properties in a node's local neighbourhood and increase the expressive power of existing GNNs. Our approach for graph feature generation consists of 3 key steps, (i) neighbourhood identification for all nodes using random walks, (ii) visualizing the neighbourhood using minimum energy approach, and (iii) training image model using the generated visualizations for node-wise feature generation. The random walk helps capturing small differences in the neighbourhood of a node by adjusting its depth and breadth parameters. While a walk captures a node's neighbourhood, it does not provide any meaningful information about the approximate length of the path between nodes. In the proposed approach, the length between nodes are preserved using a energy-based graph visualization. This can give nodes with similar neighbourhoods a different identity based on their distance from their neighbours. These structure preserving information produced by the energy based graph visualization is encoded appropriately by an image model in the final step. These encodings can then be used as input to GNN for different downstream tasks. We have shown experimentally that it is sufficient to use only 2 layers of GNN to produce state of the art results. This in-turn reduce the problem of over-smoothing without sacrificing the advantages of neighbourhood aggregation. In summary our contributions are as follows:
\begin{itemize}
    \item A novel methodology for node-feature generation of graphs that inherently retains it structural information.
    \item Present experimental evidence that the features generated by our proposed method increases the performance of existing GNN models.
    \item Reduce over-smoothing problem of GNN by only using 2 layers with the proposed features while improving the accuracy of the state of the art GNN models.
    \item Reduce the parameter space of the GNNs by using proposed embeddings that require no further training. 
    The proposed methodology use batched approach allowing scalability.
\end{itemize}
\section{GraphViz2Vec}
 \begin{figure*}
  \centering
  \includegraphics[width=\textwidth]{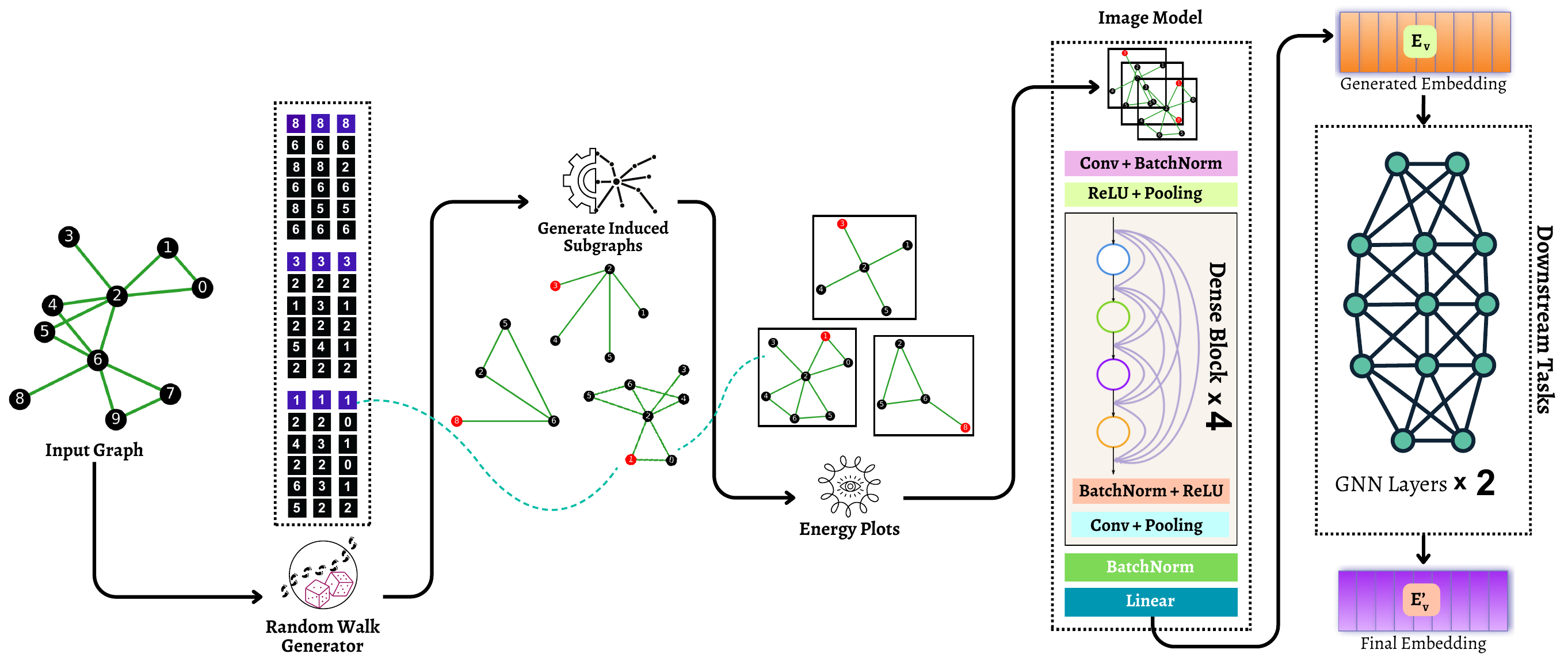}
  \caption{Proposed Model.}
  \label{fig:fig5}
 \end{figure*}
The proposed GraphViz2Vec model generates node-level features represented in vector format. These features essentially preserve the structural properties of the network via visualization of the local neighbourhood of nodes.
There are three components of the model. First, local neighbourhoods of all nodes are identified using random walks.Then various visualizations of induced subgraphs are generated for each node, using an energy-based force-directed algorithm. Finally, the generated images are trained using an image model to get the output embeddings. Given a graph $G = (V, E)$, we define the embedding of a node $v$ generated by GraphViz2Vec (represented here as $f$) as $\vec{E}_v = f(v; E, \phi, \theta)$. Here $E \in G$, $\phi$ and $\theta$ are the parameters of the function $f$. Each parameter determines the outputs of different components of the embedding pipeline. The parameter $\phi$ controls the random walk, and $\theta$ is learned from an image model. 
The block diagram of Figure \ref{fig:fig5} shows the working steps of GraphViz2Vec with example outputs of each components. The green dashed line shows the correspondence of random walk generated for node $1$ with the induced subgraph and the graph visualization. One may observe from these correspondence that the minmimum energy diagram enforced discipline in the image which is otherwise not enforced by the graph data structures. In other words, node position is not important in graph whereas the generated visualization with minimum energy ensures positional alignments. This observation is the basis of the proposed feature generation methodology. The following Sections explain these components and their parameters in more detail.


\subsection{Projection of node neighbourhood into subgraphs}
\textit{The neighbourhood of any node contains information about its local interactions and the interactions among its neighbours. These interactions form the basis for many network properties that should be implicitly present in the final encoding of a node.}

We execute multiple random walks of the same length starting from each node in the graph $G$. The probability of the depth and breadth-wise traversal of the random walk is parameterized by $d$ and $b$ and length of the random walk and the number of walks starting per node are parameterized by $l$ and $k$, respectively. We consider the subgraph induced by the union of the nodes traverse by all the random walks starting from the node as its local neighbourhood. We can set a bound on the maximum number of nodes per subgraph. We repeat the process to find multiple subgraphs for each node. The number of subgraphs for a node is parameterized by $n$. All the parameters of the random walk are unifiedly represented as $\phi$ i.e. $\phi = (d, b, l, k, n)$. The lines $3$-$14$ in the Algorithm \ref{alg1} show the steps for subgraph generation.

Random walk-based methods like DeepWalk \cite{10.1145/2623330.2623732} and Node2Vec \cite{10.1145/2939672.2939754} consider a multi-order neighbourhood structure of the nodes of a graph to generate node embeddings, thus showing that neighbourhood information plays a vital role in the quality of output embeddings. Therefore, we use random walks to extract the neighbourhood structure from a node. There are methods \cite{10.1145/2736277.2741093} that use other learnable objective functions to encode the first-order and second-order graph neighbourhood, but they take more time. In the case of large graphs, the walks can be extracted efficiently using \cite{lerer2019pytorch}, and \cite{10.1145/3308558.3313508}.

\subsection{Energy Plot of the subgraphs}
\textit{Usually, graph embedding techniques like GNNs cannot retain a node's complete structural information, making them less expressive. Much of this information is very implicit in the pictorial representation of a node's local neighbourhood. As shown in Figure \ref{fig:fig5}, this visual representation implicitly contains information such as the approximate number of neighbours, triangle participation, and degree of neighbours, which are not apparent from the adjacency matrix representation used in GNNs. These simple measures form the basis of more complex local information like degree centrality, closeness centrality, clustering and other metrics that are inherently present in the picture.} In this component minimum energy based visualization technique is used as stated below for preserving the structural properties.

The sub-graphs extracted in the previous component are pictorially represented using the Kamada-Kawai (KK) \cite{kamada1989algorithm} energy-based algorithm. 
Traditionally force-directed methods are used for visualizing graphs \cite{6295853, 7557906, 6658169, 6902873}. Recent development of graph visualizing also used GNNs \cite{9810169}. We have selected the Kamadi-Kawai \cite{kamada1989algorithm} algorithm as it considers the coordinates of the nodes in its energy equation; thus, it produces consistent results across multiple runs on the same set of nodes with minimum time complexity. In the KK algorithm, $|V|$ nodes connected with springs try to get into a balanced state with minimum energy in the springs. A higher imbalance in the spring system indicates a high-energy state. The goal is to minimize the energy state with minimum force between the springs. In the process, the algorithm brings adjacent vertices close to one another and moves non-adjacent vertices far from one another. We express the energy in the system using Equation \ref{eqn1}.

\begin{equation}
    \label{eqn1}
    \resizebox{0.9\hsize}{!}{$
    \Delta=\sum_{i=1}^{|V|-1} \sum_{j=i+1}^{|V|} \frac{1}{2} \delta_{i j}\left(\sqrt{\left(h_i-h_j\right)^2+\left(v_i-v_j\right)^2} - \tau_{i j}\right)^2$
    }
\end{equation}
In the equation, $(h_i, v_i)$ and $(h_j, v_j)$ are the $x$ and $y$ coordinates of the node in the Euclidean space. The variable $\tau_{ij} = D * p_{ij}$ where $D$ represents the length of an edge in the display pane and $p_{ij}$ is the path length between nodes $i$ and $j$ in the graph. Here, $D = D_0/\lambda$, where $D_0$ is the length of a side of the square display pane, and $\lambda$ is the graph's diameter. Including $\tau_{ij}$ as a parameter helps reduce edge crossing, creating a clear pictorial representation. The variable $\delta_{ij}$ is the ratio of the Euclidean distance to the squared path distance between two nodes, $i$ and $j$, i.e. $\delta_{ij} = \sigma/(p_{ij})^2$, where $\sigma$ is a hyperparameter. The parameter $\delta_{ij}$ provides a sense of the distance between nodes $i$ and $j$ in the plot, given their path length in the graph. Including the path length between two nodes helps encode spatial information into an image. The algorithm also retains densely connected areas in the neighbourhood of a node, indicating its density and importance in the traversal path between other nodes.  

The optimization challenge here is to find the values of the $2|V|$ variables $(h_1, h_2, ..., h_{|V|}, v_1, v_2, ...,v_{|V|})$. A local minimum is preferred since it is challenging to find the global minimum. Using the Newton-Raphson method, we can get the local minimum of Equation \ref{eqn1} from a random initial state. The necessary condition for the local minimum can be stated in Equation \ref{eqn2}.
\begin{equation}
    \label{eqn2}
    \frac{\partial \Delta}{\partial h_m}=\frac{\partial \Delta}{\partial v_m}=0 \quad \text { for } 1 \leqslant m \leqslant |V|
\end{equation}
\begin{equation}
    \label{eqn3}
    \frac{\partial \Delta}{\partial h_m}=\sum_{i \neq m} \delta_{m i}\left\{\left(h_m-h_i\right)-\frac{\tau_{m i}\left(h_m-h_i\right)}{\mathcal{E}[(h_m,v_m),(h_i, v_i)]}\right\}
\end{equation}
\begin{equation}
    \label{eqn4}
    \frac{\partial \Delta}{\partial v_m}=\sum_{i \neq m} \delta_{m i}\left\{\left(v_m-v_i\right)-\frac{\tau_{m i}\left(v_m-v_i\right)}{\mathcal{E}[(h_m,v_m),(h_i, v_i)]}\right\}
\end{equation}
In the equations \ref{eqn3} and \ref{eqn4}, $\mathcal{E}[c_i, c_m]$ represents the Euclidean distance between co-ordinates $c_i$ and $c_m$. The parameters that satisfy Equation \ref{eqn2} represent a state in which all the forces on all the springs are balanced. The $2|V|$ equations corresponding to each $h_i$ and $v_i$ are dependent on each other and, therefore, cannot be solved using a 2|V|-dimensional Newton-Raphson method. The equations are solved by considering only one particle (node) to be mobile at a time. Let us suppose that the coordinates for this particle are represented by $c_i = (h_i, v_i)$. The particle $c_i$ is moved to its stable position while keeping all the other particles frozen. This allows the authors of \cite{kamada1989algorithm} to determine the solution of $\Delta$ using a 2-dimensional Netwon-Raphson method. To reduce the number of parameters in Equations \ref{eqn2}-\ref{eqn4}, we generally keep a maximum of 256 nodes in a subgraph that gives good results without compromising performance. The absence of learning parameters makes the process significantly faster. The lines $16$-$19$ in Algorithm \ref{alg1} show the steps for generating energy plots.

\subsection{Training Image Model}
\textit{The pictorial representation contains innate structural information, which must be represented in a single-dimensional vector representation for input to a GNN for different downstream classification tasks on graphs.}

The inherent structural information in the pictorial representations is extracted using an image classification model parameterized by $\theta$. The plots generated from the subgraphs for each node are used as input to the image model. The model needs to represent the interaction of the concerned node with its surroundings. To help the image model with the task, we add node numbers to each node in the plot and change the colour of the concerned node whose neighbourhood we inspect. We split the subgraph plots into two sets, $S_1$ and $S_2$ such that $\{f_2(s)|{\forall s \in S_1}\} \cap \{f_2(s)|{\forall s \in S_2}\} = \emptyset$. The node corresponding to subgraph $s$ is represented by $f_2(s)$. We use the set $S_1$ for training and $S_2$ for testing as represented in Algorithm \ref{alg1}. The image model  is then trained to classify the nodes based on available class labels for each node. We have used different image models, but we get the best results with the DenseNet \cite{8099726} model. The model has dense blocks, with each layer receiving information from all the previous layers in the block. The dense connections help to tackle the loss of information due to the vanishing gradient problem, thus improving performance over models like Resnet. The primary information of our images is widely spread clusters with white space in between. This makes our images more perceptible to loss of information, thus explaining the better performance of the DenseNet model over others.

After training, we extract $1024$ features of shape $7 \times 7$ from the penultimate batch normalization layer of the DenseNet model for each image of a node. We apply the ReLU activation function on the extracted features followed by a two-dimensional adaptive average pooling to reduce the dimensions of the features to $1024 \times 1 \times 1$. We flatten the features to get the final output features of an image. We have taken multiple subgraphs for each node to make the image model invariant to the physical rotations of the neighbours of a node in the plot. Multiple subgraphs for a node also accurately capture the neighbourhood of a node, especially in the case of large graphs. The extracted features for all subgraphs of a node are aggregated. In this case, we use summation as the aggregation function. The initial features of the test nodes are computed by aggregating the features of their neighbours selected for training. The test nodes with no neighbours are given an encoding from the trained image model. The lines $20$-$53$ in the Algorithm \ref{alg1} show the steps for training an image model and extracting node wise features from the model.

\begin{remark}
Each of the above mentioned operations is sequentially executed in the order followed above. We save the subgraphs, followed by generating the plots for each subgraph. The generated plots are used for node feature generation using an image model, and the model with the best training accuracy is saved. The node features are extracted and saved using the best-performing model. The saved features are then input to the GNN for the specified task. The decoupling of the feature generation and GNN phase makes it suitable for generating new node features on the fly, as in the case of dynamic graphs. When a new node is attached to a graph, we get the subgraph for its local neighbourhood, using which we can generate the plots for the image model. The image model can be used to extract the features of the node from the neighbourhood plots. The trained image model can be finetuned after $t$ timesteps  using the newly added nodes to maintain the quality of generated features.
\end{remark}
\begin{algorithm}[]
\caption{GraphViz2Vec}
\label{alg1}
\begin{algorithmic}[1]
\STATE \textbf{INPUT}: Graph (G) =$(V,E)$, $d$, $b$, $k$, $l$, $n$, $size_{plot}$, $max_{nodes}$
\STATE \textbf{OUTPUT}: Node Embeddings $E_v$
\STATE $subgraphs \gets []$
\FOR{$i \in V$}
    \STATE $s = 0$
    \WHILE{$s < n$}
        \STATE $walk \gets random_{walk}(i, d, b, k, l)$
        \STATE $nodes \gets \bigcup walk$ \COMMENT{Union of all nodes in the walk}
        \IF{$|nodes| < max_{nodes}$}
            \STATE $G_1 \gets subgraph(nodes)$ \COMMENT{Create an induced subgraph}
            \STATE $subgraphs.add(G_1)$
        \ENDIF
        \STATE $s = s + 1$
    \ENDWHILE
\ENDFOR
\STATE $plots \gets []$
\FOR{$s \in subgraphs$}
    \STATE $plots.add(plot(s, size_{plot}))$ 
    \COMMENT{Make a plot of each subgraph according to Equation \ref{eqn1}}
\ENDFOR
\STATE $S_1, S_2 \gets split(plots)$ \COMMENT{$S_1$ for training, $S_2$ for testing}
\WHILE{$e < epochs$}
    \FOR{$p \in S_1$}
        \STATE $train(model_{img}, p)$
    \ENDFOR
    \STATE $e \gets e + 1$
\ENDWHILE
\STATE $features \gets \{\}$
\FOR{$p \in S_1$}
        \STATE $node \gets f_2(p)$ \COMMENT{$f_2$ returns node corresponding to plot p}
        \STATE $feat \gets model_{img}(p)$ \COMMENT{Extract features from image model}
        \IF{$features[node]$ exists}
            \STATE $features[node] \gets feat + features[node]$ 
        \ELSE
            \STATE $features[node] \gets feat$
        \ENDIF
\ENDFOR
\STATE $train_{nodes} = \{f_2(i) | i \in S_1\}$
\FOR{$p \in S_2$}
        \STATE $node \gets f_2(p)$ \COMMENT{node corresponding to plot p}
        \STATE $feat \gets model_{img}(p)$
        \STATE $neighbors_{train} \gets \{i|(i \leftrightarrow node) \in E, i \in train_{nodes}\}$
        \IF{$|neighbors_{train}| > 0$}
            \STATE $features[node] \gets \vec{0}$
            \FOR{$i \in neighbors_{train}$}
                \STATE $features[node] \gets features[node] + features[i]$ 
            \ENDFOR
        \ELSE
            \IF{$features[node]$ does not exist}
                \STATE $features[node] \gets \vec{0}$
            \ENDIF
            \STATE $features[node] \gets feature[node] + feat$
        \ENDIF
\ENDFOR
\end{algorithmic}
\end{algorithm} 
\section{Applications}
The proposed GraphViz2Vec can be used in many downstream classification tasks. In the present paper we report results of two well known classification tasks of network science, namely, node classification and link classification. Let us first define the task in the current context before presenting the experiments and results.

\begin{definition}[Node Classification]
Given a graph $G(V, E)$, the node classification problem is to predict the class of each node $i$ as $class_i = f_3(i; \theta_2, \theta_E)~\forall i \in V$. Here, $\theta_2$ is the parameter of the GNN, and $\theta_E$ represents the embeddings of the nodes. 
\end{definition}

\begin{definition}[Link Classification]
Given a graph $G(V, E)$, the link classification problem is to predict the class of each edge $(i,j)$ as $class_{ij} = f_4((i,j); \theta_3, \theta_E) ~\forall (i,j) \in E$. Here, $\theta_3$ is the parameter of the GNN, and $\theta_E$ represents the embeddings of the nodes. 
\end{definition}

If the initial encodings of all nodes are coming from GraphViz2Vec then $\theta_E = \vec{E}_v$. One should note that once we use $\vec{E}_v$ as the features no further training of embeddings is required. 
\subsection{Experiments}
\subsubsection{Datasets, Baseline and Evaluation Metrics}
We use various datasets to analyse the features generated using the proposed method. These datasets include citation networks, computer and photo networks, social networks of developers and gamers. The Table \ref{tab:table1} shows the basic statistics of these datasets. We use accuracy as the evaluation metric for both node and link classification tasks.

We have used our generated features with 12 existing GNN models \cite{10225480, 10.1609/aaai.v33i01.33014602, veličković2018graph, bresson2018residual, brody2022how, 10.5555/3524938.3525045, ijcai2021p214, 10.5555/3495724.3497151, izadi2020optimization} including classical GCN and SAGE models \cite{kipf2017semisupervised, 10.5555/3294771.3294869}. The SSP \cite{izadi2020optimization} model performs node classification and uses one of the abovementioned models in the backend, along with certain refinements like optimization using natural gradients. Therefore, we have only compared the results of SSP for the datasets they have reported on to highlight the improvement by the generated features. The TransSage model is a combination model of graph transformer \cite{ijcai2021p214} and SAGE Convolution layer.
\begin{table}[h]
\caption{The basic statistics of Datasets}
\label{tab:table1}
\resizebox{\columnwidth}{!}{%
\begin{tabular}{|r|r|r|r|r|r|r|r|r|}
\hline
\multicolumn{1}{|c|}{\textbf{Datasets}} & \multicolumn{1}{c|}{\textbf{Nodes}} & \multicolumn{1}{c|}{\textbf{Edges}} & \multicolumn{1}{c|}{\textbf{Density}} & \multicolumn{1}{c|}{\textbf{\begin{tabular}[c]{@{}c@{}}Avg \\ Deg\\ /Node\end{tabular}}} & \multicolumn{1}{c|}{\textbf{\begin{tabular}[c]{@{}c@{}}Avg \\ Neighbor\\ Deg/Node\end{tabular}}} & \multicolumn{1}{c|}{\textbf{\begin{tabular}[c]{@{}c@{}}Avg \\ Triangle\\ /Node\end{tabular}}} & \multicolumn{1}{c|}{\textbf{Class}} & \multicolumn{1}{c|}{\textbf{\begin{tabular}[c]{@{}c@{}}Feature\\ Size\end{tabular}}} \\ \hline
Cora \cite{pmlr-v48-yanga16} & 2708 & 10556 & 0.00144 & 7.796 & 11.631 & 1.805 & 7 & 1433 \\ \hline
CiteSeer \cite{pmlr-v48-yanga16} & 3327 & 9104 & 0.00082 & 5.4727 & 5.4973 & 1.052 & 6 & 3703 \\ \hline
PubMed \cite{pmlr-v48-yanga16} & 19717 & 88648 & 0.00023 & 8.9920 & 19.2743 & 1.904 & 3 & 500 \\ \hline
Photo \cite{shchur2019pitfalls}& 7650 & 238162 & 0.00407 & 62.264 & 101.656 & 281.331 & 8 & 745 \\ \hline
Computers \cite{shchur2019pitfalls} & 13752 & 491722 & 0.0026 & 71.512 & 163.138 & 333.217 & 10 & 767 \\ \hline
ES \cite{9514682}& 4648 & 123412 & 0.00571 & 53.103 & 146.389 & 129.180 & 2 & 128 \\ \hline
FR \cite{9514682}& 6551 & 231883 & 0.0054 & 70.793 & 276.133 & 193.570 & 2 & 128 \\ \hline
PT \cite{9514682}& 1912 & 64510 & 0.01766 & 67.479 & 152.153 & 272.243 & 2 & 128 \\ \hline
GitHub\cite{9514682} & 37700 & 578006 & 0.00041 & 30.663 & 818.149 & 41.682 & 2 & 0 \\ \hline
Flickr \cite{Zeng2020GraphSAINT:}& 89250 & 899756 & 0.00011 & 20.162 & 156.116 & 2.1492 & 7 & 500 \\ \hline
\end{tabular}%
}
\end{table}

\subsubsection{Experiment Setting}
We have compared the performance of our generated features with the feature set present with the datasets. We also compare our features with trainable random initial embeddings for a node. All the models used for comparison have $2$ layers, reducing the over-smoothing problem. We have experimented with various sizes for the initial embeddings and reported the best results for the node classification and link classification tasks in Table \ref{tab:table2} and \ref{tab:table3}, respectively. The `Actual' row in the Tables is the performance of the existing models on the feature set provided with the datasets; the `Generated' row is the performance of the models with our feature set. The `Generated + Actual' row shows the performance of the models when we concatenate the generated and actual features. We do not use the `Generated + Actual' row for link classification as it shows no significant improvements. The results shown in Table \ref{tab:table2} and \ref{tab:table3} are averaged over multiple runs. We use multiple random walks of length 128 for the CiteSeer dataset, length 32 for the Cora, Pubmed, Photo, Computers, ES, and FR datasets and length 64 for the GitHub and Flickr datasets. We have taken shorter random walks for each node of denser graphs to reduce edge crossing in the plot. We select the length of the random walks and the number of walks per node based on visual inspection of a small set of sample nodes for each dataset. The point of observation was to keep similar subgraphs for neighbouring nodes and dissimilar subgraphs for non-neighbouring nodes. We increase the number of subgraphs per node in scenarios where the walk produces high-variance neighbourhood subgraphs for a node. Increasing the number of subgraphs (plots) also helps the model to be more invariant to the arrangement of the nodes in a 2-D space. We show how the size of the random walk affects the model performance in an ablation study. We have performed the experiments in a single A30 GPU with a 64 core CPU and 256 GB of RAM.

\begin{table*}[h]
\caption{Node classification task: Values represent model's accuracy.}
\label{tab:table2}
\resizebox{0.8\hsize}{!}{%
\begin{tabular}{llrrrrrrrrrrrcr}
\hline
                                             &                                      & \multicolumn{1}{l}{M1}                & \multicolumn{1}{l}{M2}                & \multicolumn{1}{l}{M3}                & \multicolumn{1}{l}{M4}                & \multicolumn{1}{l}{M5}                & \multicolumn{1}{l}{M6}                & \multicolumn{1}{l}{M7}                & \multicolumn{1}{l}{M8}                & \multicolumn{1}{l}{M9}                & \multicolumn{1}{l}{M10}               & \multicolumn{1}{l}{M11}               & \multicolumn{1}{l}{M12}                                   & \multicolumn{1}{l}{Max Acc.}          \\ \hline
                                             & Actual                               & 0.87                                  & 0.86                                  & 0.85                                  & 0.84                                  & 0.86                                  & 0.85                                  & 0.88                                  & 0.81                                  & 0.85                                  & 0.87                                  & 0.85                                  & \multicolumn{1}{l}{0.90}                                  & 0.90                                  \\
                                             & \cellcolor[HTML]{EFEFEF}Generated    & \cellcolor[HTML]{EFEFEF}0.86          & \cellcolor[HTML]{EFEFEF}0.86          & \cellcolor[HTML]{EFEFEF}0.87          & \cellcolor[HTML]{EFEFEF}0.86          & \cellcolor[HTML]{EFEFEF}0.87          & \cellcolor[HTML]{EFEFEF}0.86          & \cellcolor[HTML]{EFEFEF}0.87          & \cellcolor[HTML]{EFEFEF}\textbf{0.85} & \cellcolor[HTML]{EFEFEF}0.87          & \cellcolor[HTML]{EFEFEF}0.84          & \cellcolor[HTML]{EFEFEF}0.86          & \multicolumn{1}{l}{\cellcolor[HTML]{EFEFEF}\textbf{0.92}} & \cellcolor[HTML]{EFEFEF}\textbf{0.92} \\
                                             & \cellcolor[HTML]{EFEFEF}Gen + Actual & \cellcolor[HTML]{EFEFEF}\textbf{0.89} & \cellcolor[HTML]{EFEFEF}\textbf{0.89} & \cellcolor[HTML]{EFEFEF}\textbf{0.88} & \cellcolor[HTML]{EFEFEF}\textbf{0.89} & \cellcolor[HTML]{EFEFEF}\textbf{0.88} & \cellcolor[HTML]{EFEFEF}\textbf{0.89} & \cellcolor[HTML]{EFEFEF}\textbf{0.89} & \cellcolor[HTML]{EFEFEF}0.84          & \cellcolor[HTML]{EFEFEF}\textbf{0.88} & \cellcolor[HTML]{EFEFEF}\textbf{0.88} & \cellcolor[HTML]{EFEFEF}\textbf{0.90} & \cellcolor[HTML]{EFEFEF}-                                 & \cellcolor[HTML]{EFEFEF}0.90          \\
\multirow{-4}{*}{Cora}                       & Random                               & 0.81                                  & 0.83                                  & 0.85                                  & 0.84                                  & 0.82                                  & 0.84                                  & 0.81                                  & 0.73                                  & 0.84                                  & 0.83                                  & 0.84                                  & -                                                         & 0.85                                  \\ \hline
                                             & Actual                               & 0.79                                  & 0.79                                  & 0.75                                  & 0.79                                  & 0.77                                  & 0.77                                  & 0.78                                  & 0.73                                  & 0.77                                  & 0.67                                  & 0.78                                  & \multicolumn{1}{l}{\textbf{0.80}}                         & 0.80                                  \\
                                             & \cellcolor[HTML]{EFEFEF}Generated    & \cellcolor[HTML]{EFEFEF}\textbf{0.89} & \cellcolor[HTML]{EFEFEF}\textbf{0.96} & \cellcolor[HTML]{EFEFEF}\textbf{0.98} & \cellcolor[HTML]{EFEFEF}\textbf{0.88} & \cellcolor[HTML]{EFEFEF}\textbf{0.98} & \cellcolor[HTML]{EFEFEF}\textbf{0.88} & \cellcolor[HTML]{EFEFEF}\textbf{0.98} & \cellcolor[HTML]{EFEFEF}\textbf{0.97} & \cellcolor[HTML]{EFEFEF}\textbf{0.93} & \cellcolor[HTML]{EFEFEF}\textbf{0.97} & \cellcolor[HTML]{EFEFEF}\textbf{0.99} & \multicolumn{1}{l}{\cellcolor[HTML]{EFEFEF}\textbf{0.89}} & \cellcolor[HTML]{EFEFEF}\textbf{0.99} \\
\multirow{-3}{*}{CiteSeer}                   & Random                               & 0.71                                  & 0.72                                  & 0.70                                  & 0.70                                  & 0.70                                  & 0.66                                  & 0.57                                  & 0.67                                  & 0.72                                  & 0.32                                  & 0.72                                  & \multicolumn{1}{l}{-}                                     & 0.72                                  \\ \hline
                                             & Actual                               & \textbf{0.85}                         & \textbf{0.87}                         & \textbf{0.87}                         & \textbf{0.86}                         & \textbf{0.89}                         & 0.84                                  & \textbf{0.86}                         & \textbf{0.87}                         & \textbf{0.87}                         & \textbf{0.90}                         & 0.86                                  & \multicolumn{1}{l}{\textbf{0.89}}                         & \textbf{0.89}                         \\
                                             & \cellcolor[HTML]{EFEFEF}Generated    & \cellcolor[HTML]{EFEFEF}0.84          & \cellcolor[HTML]{EFEFEF}0.83          & \cellcolor[HTML]{EFEFEF}0.84          & \cellcolor[HTML]{EFEFEF}0.84          & \cellcolor[HTML]{EFEFEF}0.84          & \cellcolor[HTML]{EFEFEF}0.84          & \cellcolor[HTML]{EFEFEF}0.83          & \cellcolor[HTML]{EFEFEF}0.63          & \cellcolor[HTML]{EFEFEF}0.84          & \cellcolor[HTML]{EFEFEF}0.84          & \cellcolor[HTML]{EFEFEF}0.85          & \multicolumn{1}{l}{\cellcolor[HTML]{EFEFEF}0.86}          & \cellcolor[HTML]{EFEFEF}0.86          \\
                                             & \cellcolor[HTML]{EFEFEF}Gen + Actual & \cellcolor[HTML]{EFEFEF}\textbf{0.85} & \cellcolor[HTML]{EFEFEF}0.85          & \cellcolor[HTML]{EFEFEF}0.85          & \cellcolor[HTML]{EFEFEF}0.85          & \cellcolor[HTML]{EFEFEF}0.85          & \cellcolor[HTML]{EFEFEF}\textbf{0.85} & \cellcolor[HTML]{EFEFEF}0.84          & \cellcolor[HTML]{EFEFEF}0.84          & \cellcolor[HTML]{EFEFEF}0.85          & \cellcolor[HTML]{EFEFEF}0.84          & \cellcolor[HTML]{EFEFEF}\textbf{0.87} & \cellcolor[HTML]{EFEFEF}-                                 & \cellcolor[HTML]{EFEFEF}0.87          \\
\multirow{-4}{*}{PubMed}                     & Random                               & 0.69                                  & 0.80                                  & 0.78                                  & 0.81                                  & 0.78                                  & 0.79                                  & 0.81                                  & 0.63                                  & 0.77                                  & 0.75                                  & 0.80                                  & -                                                         & 0.81                                  \\ \hline
                                             & Actual                               & 0.82                                  & 0.89                                  & 0.90                                  & 0.85                                  & 0.90                                  & 0.86                                  & \textbf{0.92}                         & 0.92                                  & 0.87                                  & 0.87                                  & 0.90                                  & -                                                         & 0.92                                  \\
                                             & \cellcolor[HTML]{EFEFEF}Generated    & \cellcolor[HTML]{EFEFEF}\textbf{0.89} & \cellcolor[HTML]{EFEFEF}0.91          & \cellcolor[HTML]{EFEFEF}\textbf{0.91} & \cellcolor[HTML]{EFEFEF}\textbf{0.92} & \cellcolor[HTML]{EFEFEF}0.90          & \cellcolor[HTML]{EFEFEF}\textbf{0.92} & \cellcolor[HTML]{EFEFEF}0.91          & \cellcolor[HTML]{EFEFEF}0.90          & \cellcolor[HTML]{EFEFEF}0.52          & \cellcolor[HTML]{EFEFEF}0.88          & \cellcolor[HTML]{EFEFEF}0.92          & \cellcolor[HTML]{EFEFEF}-                                 & \cellcolor[HTML]{EFEFEF}0.92          \\
                                             & \cellcolor[HTML]{EFEFEF}Gen + Actual & \cellcolor[HTML]{EFEFEF}\textbf{0.89} & \cellcolor[HTML]{EFEFEF}\textbf{0.92} & \cellcolor[HTML]{EFEFEF}0.90          & \cellcolor[HTML]{EFEFEF}\textbf{0.92} & \cellcolor[HTML]{EFEFEF}\textbf{0.92} & \cellcolor[HTML]{EFEFEF}\textbf{0.92} & \cellcolor[HTML]{EFEFEF}\textbf{0.92} & \cellcolor[HTML]{EFEFEF}\textbf{0.94} & \cellcolor[HTML]{EFEFEF}0.88          & \cellcolor[HTML]{EFEFEF}0.89          & \cellcolor[HTML]{EFEFEF}\textbf{0.93} & \cellcolor[HTML]{EFEFEF}-                                 & \cellcolor[HTML]{EFEFEF}\textbf{0.94} \\
\multirow{-4}{*}{Photo}                      & Random                               & \textbf{0.89}                         & 0.91                                  & \textbf{0.91}                         & 0.90                                  & 0.90                                  & 0.90                                  & 0.89                                  & 0.89                                  & \textbf{0.91}                         & \textbf{0.91}                         & 0.89                                  & -                                                         & 0.91                                  \\ \hline
                                             & Actual                               & 0.65                                  & 0.75                                  & 0.59                                  & 0.71                                  & 0.77                                  & 0.68                                  & 0.73                                  & 0.52                                  & 0.57                                  & 0.78                                  & 0.76                                  & -                                                         & 0.78                                  \\
                                             & \cellcolor[HTML]{EFEFEF}Generated    & \cellcolor[HTML]{EFEFEF}0.81          & \cellcolor[HTML]{EFEFEF}\textbf{0.87} & \cellcolor[HTML]{EFEFEF}\textbf{0.82} & \cellcolor[HTML]{EFEFEF}0.81          & \cellcolor[HTML]{EFEFEF}\textbf{0.88} & \cellcolor[HTML]{EFEFEF}0.81          & \cellcolor[HTML]{EFEFEF}\textbf{0.89} & \cellcolor[HTML]{EFEFEF}\textbf{0.77} & \cellcolor[HTML]{EFEFEF}0.77          & \cellcolor[HTML]{EFEFEF}\textbf{0.84} & \cellcolor[HTML]{EFEFEF}\textbf{0.88} & \cellcolor[HTML]{EFEFEF}-                                 & \cellcolor[HTML]{EFEFEF}\textbf{0.89} \\
\multirow{-3}{*}{Computers}                  & Random                               & \textbf{0.83}                         & 0.85                                  & 0.82                                  & \textbf{0.84}                         & 0.86                                  & \textbf{0.86}                         & 0.84                                  & 0.76                                  & \textbf{0.84}                         & 0.63                                  & 0.87                                  & -                                                         & 0.86                                  \\ \hline
\multicolumn{1}{c}{}                         & Actual                               & 0.70                                  & \textbf{0.70}                         & 0.70                                  & 0.70                                  & 0.70                                  & 0.70                                  & 0.70                                  & 0.65                                  & \textbf{0.70}                         & \textbf{0.70}                         & 0.70                                  & -                                                         & 0.70                                  \\
\multicolumn{1}{c}{}                         & \cellcolor[HTML]{EFEFEF}Generated    & \cellcolor[HTML]{EFEFEF}\textbf{0.73} & \cellcolor[HTML]{EFEFEF}\textbf{0.70} & \cellcolor[HTML]{EFEFEF}\textbf{0.72} & \cellcolor[HTML]{EFEFEF}\textbf{0.71} & \cellcolor[HTML]{EFEFEF}\textbf{0.72} & \cellcolor[HTML]{EFEFEF}\textbf{0.71} & \cellcolor[HTML]{EFEFEF}\textbf{0.71} & \cellcolor[HTML]{EFEFEF}\textbf{0.71} & \cellcolor[HTML]{EFEFEF}\textbf{0.70} & \cellcolor[HTML]{EFEFEF}\textbf{0.70} & \cellcolor[HTML]{EFEFEF}\textbf{0.74} & \cellcolor[HTML]{EFEFEF}-                                 & \cellcolor[HTML]{EFEFEF}\textbf{0.74} \\
\multicolumn{1}{c}{\multirow{-3}{*}{ES}}     & Random                               & 0.65                                  & 0.68                                  & 0.69                                  & 0.65                                  & 0.71                                  & 0.64                                  & 0.62                                  & 0.63                                  & 0.69                                  & 0.67                                  & 0.68                                  & -                                                         & 0.71                                  \\ \hline
\multicolumn{1}{c}{}                         & Actual                               & \textbf{0.63}                         & \textbf{0.63}                         & \textbf{0.63}                         & \textbf{0.63}                         & \textbf{0.63}                         & \textbf{0.63}                         & \textbf{0.63}                         & 0.59                                  & \textbf{0.63}                         & 0.62                                  & 0.62                                  & -                                                         & 0.63                                  \\
\multicolumn{1}{c}{}                         & \cellcolor[HTML]{EFEFEF}Generated    & \cellcolor[HTML]{EFEFEF}\textbf{0.63} & \cellcolor[HTML]{EFEFEF}\textbf{0.63} & \cellcolor[HTML]{EFEFEF}\textbf{0.63} & \cellcolor[HTML]{EFEFEF}0.61          & \cellcolor[HTML]{EFEFEF}0.59          & \cellcolor[HTML]{EFEFEF}\textbf{0.63} & \cellcolor[HTML]{EFEFEF}0.62          & \cellcolor[HTML]{EFEFEF}0.62          & \cellcolor[HTML]{EFEFEF}\textbf{0.63} & \cellcolor[HTML]{EFEFEF}\textbf{0.63} & \cellcolor[HTML]{EFEFEF}\textbf{0.65} & \cellcolor[HTML]{EFEFEF}-                                 & \cellcolor[HTML]{EFEFEF}\textbf{0.65} \\
\multicolumn{1}{c}{}                         & \cellcolor[HTML]{EFEFEF}Gen + Actual & \cellcolor[HTML]{EFEFEF}\textbf{0.63} & \cellcolor[HTML]{EFEFEF}\textbf{0.63} & \cellcolor[HTML]{EFEFEF}\textbf{0.63} & \cellcolor[HTML]{EFEFEF}\textbf{0.63} & \cellcolor[HTML]{EFEFEF}0.62          & \cellcolor[HTML]{EFEFEF}\textbf{0.63} & \cellcolor[HTML]{EFEFEF}\textbf{0.63} & \cellcolor[HTML]{EFEFEF}\textbf{0.63} & \cellcolor[HTML]{EFEFEF}\textbf{0.63} & \cellcolor[HTML]{EFEFEF}0.62          & \cellcolor[HTML]{EFEFEF}0.65          & \cellcolor[HTML]{EFEFEF}-                                 & \cellcolor[HTML]{EFEFEF}\textbf{0.65} \\
\multicolumn{1}{c}{\multirow{-4}{*}{FR}}     & Random                               & 0.60                                  & 0.56                                  & 0.58                                  & 0.60                                  & 0.57                                  & 0.58                                  & 0.53                                  & 0.54                                  & 0.57                                  & 0.58                                  & 0.58                                  & -                                                         & 0.60                                  \\ \hline
\multicolumn{1}{c}{}                         & Actual                               & 0.64                                  & \textbf{0.67}                         & 0.68                                  & 0.64                                  & 0.60                                  & 0.64                                  & 0.68                                  & 0.62                                  & 0.62                                  & 0.64                                  & 0.62                                  & -                                                         & 0.68                                  \\
\multicolumn{1}{c}{}                         & \cellcolor[HTML]{EFEFEF}Generated    & \cellcolor[HTML]{EFEFEF}\textbf{0.70} & \cellcolor[HTML]{EFEFEF}\textbf{0.67} & \cellcolor[HTML]{EFEFEF}\textbf{0.69} & \cellcolor[HTML]{EFEFEF}\textbf{0.71} & \cellcolor[HTML]{EFEFEF}\textbf{0.69} & \cellcolor[HTML]{EFEFEF}\textbf{0.69} & \cellcolor[HTML]{EFEFEF}\textbf{0.69} & \cellcolor[HTML]{EFEFEF}\textbf{0.68} & \cellcolor[HTML]{EFEFEF}\textbf{0.70} & \cellcolor[HTML]{EFEFEF}\textbf{0.68} & \cellcolor[HTML]{EFEFEF}\textbf{0.70} & \cellcolor[HTML]{EFEFEF}-                                 & \cellcolor[HTML]{EFEFEF}\textbf{0.70} \\
\multicolumn{1}{c}{\multirow{-3}{*}{PT}}     & Random                               & 0.63                                  & 0.63                                  & 0.64                                  & 0.57                                  & 0.65                                  & 0.61                                  & 0.60                                  & 0.56                                  & 0.64                                  & 0.64                                  & 0.60                                  & -                                                         & 0.64                                  \\ \hline
\multicolumn{1}{c}{}                         & Random                               & 0.75                                  & 0.80                                  & 0.78                                  & 0.80                                  & 0.77                                  & 0.81                                  & 0.79                                  & 0.74                                  & 0.79                                  & 0.74                                  & 0.76                                  & -                                                         & 0.81                                  \\
\multicolumn{1}{c}{\multirow{-2}{*}{GitHub}} & \cellcolor[HTML]{EFEFEF}Generated    & \cellcolor[HTML]{EFEFEF}\textbf{0.84} & \cellcolor[HTML]{EFEFEF}\textbf{0.83} & \cellcolor[HTML]{EFEFEF}\textbf{0.83} & \cellcolor[HTML]{EFEFEF}\textbf{0.84} & \cellcolor[HTML]{EFEFEF}\textbf{0.83} & \cellcolor[HTML]{EFEFEF}\textbf{0.84} & \cellcolor[HTML]{EFEFEF}\textbf{0.83} & \cellcolor[HTML]{EFEFEF}\textbf{0.84} & \cellcolor[HTML]{EFEFEF}\textbf{0.82} & \cellcolor[HTML]{EFEFEF}\textbf{0.82} & \cellcolor[HTML]{EFEFEF}\textbf{0.85} & \cellcolor[HTML]{EFEFEF}-                                 & \cellcolor[HTML]{EFEFEF}\textbf{0.85} \\ \hline
\multicolumn{1}{c}{}                         & Actual                               & 0.47                                  & \textbf{0.46}                         & 0.42                                  & 0.46                                  & 0.43                                  & 0.46                                  & \textbf{0.48}                         & 0.39                                  & 0.42                                  & \textbf{0.49}                         & \textbf{0.46}                         & -                                                         & 0.49                                  \\
\multicolumn{1}{c}{}                         & \cellcolor[HTML]{EFEFEF}Generated    & \cellcolor[HTML]{EFEFEF}\textbf{0.53} & \cellcolor[HTML]{EFEFEF}\textbf{0.46} & \cellcolor[HTML]{EFEFEF}0.47          & \cellcolor[HTML]{EFEFEF}\textbf{0.49} & \cellcolor[HTML]{EFEFEF}\textbf{0.45} & \cellcolor[HTML]{EFEFEF}\textbf{0.51} & \cellcolor[HTML]{EFEFEF}0.45          & \cellcolor[HTML]{EFEFEF}\textbf{0.45} & \cellcolor[HTML]{EFEFEF}\textbf{0.51} & \cellcolor[HTML]{EFEFEF}0.43          & \cellcolor[HTML]{EFEFEF}\textbf{0.46} & \cellcolor[HTML]{EFEFEF}-                                 & \cellcolor[HTML]{EFEFEF}\textbf{0.53} \\
\multicolumn{1}{c}{\multirow{-3}{*}{Flickr}} & Random                               & 0.51                                  & 0.39                                  & \textbf{0.51}                         & 0.48                                  & 0.43                                  & 0.46                                  & 0.41                                  & 0.38                                  & 0.47                                  & 0.36                                  & 0.45                                  & -                                                         & 0.51                                  \\ \hline
\multicolumn{15}{l}{\begin{tabular}[c]{@{}l@{}}*\rule{0pt}{3ex} Models M1:GCN, M2:SAGE, M3:GraphCon , M4:GAT, M5:ResGated, M6:GATv2\end{tabular}}\\ 
\multicolumn{15}{l}{\begin{tabular}[c]{@{}l@{}}*\rule{0pt}{3ex} M7:TransConv, M8:Gen, M9:General , M10:FiLM, M11:TranSage, M12:SSP\end{tabular}}\\
\end{tabular}%
}
\end{table*}

\begin{table*}[h]
\caption{Link Classification Task: The values represent the model’s accuracy in predicting positive and negative edges. \tablefootnote{Models M1-M11 are the same as Table \ref{tab:table2}}}
\label{tab:table3}
\resizebox{0.75\hsize}{!}{%
\begin{tabular}{ccrrrrrrrrrrrr}
\hline
\multicolumn{1}{l}{} & \multicolumn{1}{l}{} & \multicolumn{1}{l}{M1} & \multicolumn{1}{l}{M2} & \multicolumn{1}{l}{M3} & \multicolumn{1}{l}{M4} & \multicolumn{1}{l}{M5} & \multicolumn{1}{l}{M6} & \multicolumn{1}{l}{M7} & \multicolumn{1}{l}{M8} & \multicolumn{1}{l}{M9} & \multicolumn{1}{l}{M10} & \multicolumn{1}{l}{M11} & \multicolumn{1}{l}{Max Acc.} \\ \hline
\rowcolor[HTML]{FFFFFF} 
\cellcolor[HTML]{FFFFFF} & Actual & 0.47 & \textbf{0.55} & \textbf{0.53} & 0.47 & \textbf{0.54} & 0.49 & \textbf{0.54} & 0.46 & 0.54 & \textbf{0.55} & \textbf{0.53} & 0.55 \\
\rowcolor[HTML]{EFEFEF} 
\cellcolor[HTML]{FFFFFF} & Generated & \textbf{0.50} & 0.50 & 0.49 & \textbf{0.50} & 0.53 & \textbf{0.52} & 0.53 & \textbf{0.55} & \textbf{0.57} & 0.52 & 0.52 & \textbf{0.57} \\
\rowcolor[HTML]{FFFFFF} 
\multirow{-3}{*}{\cellcolor[HTML]{FFFFFF}Cora} & Random & \textbf{0.50} & 0.49 & 0.51 & \textbf{0.50} & 0.52 & 0.50 & 0.50 & 0.50 & 0.50 & 0.53 & 0.50 & 0.53 \\ \hline
 & Actual & 0.46 & \textbf{0.53} & 0.51 & 0.48 & 0.53 & 0.49 & 0.53 & 0.32 & 0.53 & \textbf{0.54} & 0.52 & 0.54 \\
 & \cellcolor[HTML]{EFEFEF}{\color[HTML]{000000} Generated} & \cellcolor[HTML]{EFEFEF}{\color[HTML]{000000} \textbf{0.52}} & \cellcolor[HTML]{EFEFEF}{\color[HTML]{000000} 0.50} & \cellcolor[HTML]{EFEFEF}{\color[HTML]{000000} \textbf{0.53}} & \cellcolor[HTML]{EFEFEF}{\color[HTML]{000000} \textbf{0.51}} & \cellcolor[HTML]{EFEFEF}{\color[HTML]{000000} \textbf{0.54}} & \cellcolor[HTML]{EFEFEF}{\color[HTML]{000000} \textbf{0.51}} & \cellcolor[HTML]{EFEFEF}{\color[HTML]{000000} \textbf{0.55}} & \cellcolor[HTML]{EFEFEF}{\color[HTML]{000000} \textbf{0.57}} & \cellcolor[HTML]{EFEFEF}{\color[HTML]{000000} \textbf{0.56}} & \cellcolor[HTML]{EFEFEF}{\color[HTML]{000000} \textbf{0.54}} & \cellcolor[HTML]{EFEFEF}{\color[HTML]{000000} \textbf{0.54}} & \cellcolor[HTML]{EFEFEF}{\color[HTML]{000000} \textbf{0.57}} \\
\multirow{-3}{*}{CiteSeer} & Random & 0.50 & 0.51 & 0.50 & 0.50 & 0.50 & 0.49 & 0.51 & 0.51 & 0.52 & 0.51 & 0.48 & 0.52 \\ \hline
 & Actual & 0.49 & 0.51 & \textbf{0.50} & \textbf{0.50} & \textbf{0.50} & 0.49 & \textbf{0.52} & 0.52 & 0.49 & \textbf{0.52} & 0.53 & 0.53 \\
 & \cellcolor[HTML]{EFEFEF}Generated & \cellcolor[HTML]{EFEFEF}\textbf{0.50} & \cellcolor[HTML]{EFEFEF}\textbf{0.52} & \cellcolor[HTML]{EFEFEF}\textbf{0.50} & \cellcolor[HTML]{EFEFEF}\textbf{0.50} & \cellcolor[HTML]{EFEFEF}\textbf{0.50} & \cellcolor[HTML]{EFEFEF}\textbf{0.52} & \cellcolor[HTML]{EFEFEF}\textbf{0.52} & \cellcolor[HTML]{EFEFEF}\textbf{0.59} & \cellcolor[HTML]{EFEFEF}\textbf{0.52} & \cellcolor[HTML]{EFEFEF}0.50 & \cellcolor[HTML]{EFEFEF}\textbf{0.54} & \cellcolor[HTML]{EFEFEF}\textbf{0.59} \\
\multirow{-3}{*}{PubMed} & Random & 0.49 & 0.51 & 0.51 & 0.49 & \textbf{0.50} & 0.49 & 0.51 & 0.54 & 0.50 & \textbf{0.52} & 0.50 & 0.54 \\ \hline
 & Actual & \textbf{0.46} & 0.49 & 0.49 & \textbf{0.48} & 0.50 & 0.50 & \textbf{0.50} & \textbf{0.63} & \textbf{0.63} & 0.55 & 0.50 & \textbf{0.63} \\
 & \cellcolor[HTML]{EFEFEF}Generated & \cellcolor[HTML]{EFEFEF}\textbf{0.46} & \cellcolor[HTML]{EFEFEF}\textbf{0.51} & \cellcolor[HTML]{EFEFEF}0.49 & \cellcolor[HTML]{EFEFEF}\textbf{0.48} & \cellcolor[HTML]{EFEFEF}0.50 & \cellcolor[HTML]{EFEFEF}\textbf{0.52} & \cellcolor[HTML]{EFEFEF}0.48 & \cellcolor[HTML]{EFEFEF}0.59 & \cellcolor[HTML]{EFEFEF}0.53 & \cellcolor[HTML]{EFEFEF}\textbf{0.60} & \cellcolor[HTML]{EFEFEF}\textbf{0.51} & \cellcolor[HTML]{EFEFEF}0.60 \\
\multirow{-3}{*}{ES} & Random & \textbf{0.46} & \textbf{0.51} & \textbf{0.51} & \textbf{0.48} & \textbf{0.54} & 0.47 & 0.49 & 0.58 & 0.56 & 0.55 & 0.48 & 0.58 \\ \hline
 & Actual & 0.44 & 0.48 & 0.49 & 0.48 & 0.50 & \textbf{0.48} & 0.48 & 0.51 & \textbf{0.61} & 0.55 & 0.49 & 0.61 \\
 & \cellcolor[HTML]{EFEFEF}Generated & \cellcolor[HTML]{EFEFEF}\textbf{0.46} & \cellcolor[HTML]{EFEFEF}\textbf{0.49} & \cellcolor[HTML]{EFEFEF}\textbf{0.50} & \cellcolor[HTML]{EFEFEF}\textbf{0.50} & \cellcolor[HTML]{EFEFEF}\textbf{0.55} & \cellcolor[HTML]{EFEFEF}\textbf{0.48} & \cellcolor[HTML]{EFEFEF}0.50 & \cellcolor[HTML]{EFEFEF}\textbf{0.65} & \cellcolor[HTML]{EFEFEF}0.50 & \cellcolor[HTML]{EFEFEF}\textbf{0.60} & \cellcolor[HTML]{EFEFEF}\textbf{0.50} & \cellcolor[HTML]{EFEFEF}\textbf{0.65} \\
\multirow{-3}{*}{FR} & Random & 0.44 & 0.48 & 0.49 & 0.48 & 0.51 & 0.47 & \textbf{0.51} & 0.51 & 0.56 & 0.55 & 0.48 & 0.56 \\ \hline
 & Actual & 0.44 & 0.48 & 0.49 & 0.47 & 0.50 & 0.48 & 0.48 & 0.51 & \textbf{0.61} & 0.55 & 0.49 & 0.61 \\
 & \cellcolor[HTML]{EFEFEF}Generated & \cellcolor[HTML]{EFEFEF}\textbf{0.49} & \cellcolor[HTML]{EFEFEF}0.49 & \cellcolor[HTML]{EFEFEF}\textbf{0.54} & \cellcolor[HTML]{EFEFEF}\textbf{0.48} & \cellcolor[HTML]{EFEFEF}\textbf{0.55} & \cellcolor[HTML]{EFEFEF}\textbf{0.65} & \cellcolor[HTML]{EFEFEF}\textbf{0.57} & \cellcolor[HTML]{EFEFEF}\textbf{0.60} & \cellcolor[HTML]{EFEFEF}\textbf{0.61} & \cellcolor[HTML]{EFEFEF}\textbf{0.56} & \cellcolor[HTML]{EFEFEF}0.47 & \cellcolor[HTML]{EFEFEF}\textbf{0.65} \\
\multirow{-3}{*}{PT} & Random & 0.46 & \textbf{0.50} & 0.49 & 0.44 & 0.53 & 0.45 & 0.51 & 0.55 & 0.60 & 0.51 & \textbf{0.50} & 0.60 \\ \hline
 & \multicolumn{1}{l}{Actual} & 0.49 & 0.50 & \textbf{0.48} & \textbf{0.50} & 0.49 & 0.49 & \textbf{0.51} & 0.49 & 0.49 & 0.50 & \textbf{0.51} & 0.51 \\
 & \multicolumn{1}{l}{\cellcolor[HTML]{EFEFEF}Generated} & \cellcolor[HTML]{EFEFEF}\textbf{0.50} & \cellcolor[HTML]{EFEFEF}\textbf{0.52} & \cellcolor[HTML]{EFEFEF}\textbf{0.48} & \cellcolor[HTML]{EFEFEF}\textbf{0.50} & \cellcolor[HTML]{EFEFEF}\textbf{0.53} & \cellcolor[HTML]{EFEFEF}\textbf{0.51} & \cellcolor[HTML]{EFEFEF}\textbf{0.51} & \cellcolor[HTML]{EFEFEF}\textbf{0.54} & \cellcolor[HTML]{EFEFEF}\textbf{0.50} & \cellcolor[HTML]{EFEFEF}\textbf{0.52} & \cellcolor[HTML]{EFEFEF}0.50 & \cellcolor[HTML]{EFEFEF}\textbf{0.54} \\
\multirow{-3}{*}{Computers} & \multicolumn{1}{l}{Random} & 0.47 & 0.48 & 0.46 & 0.48 & 0.44 & 0.49 & 0.46 & 0.45 & 0.47 & 0.44 & 0.48 & 0.49 \\ \hline
\multicolumn{1}{l}{} & \multicolumn{1}{l}{\cellcolor[HTML]{EFEFEF}Generated} & \cellcolor[HTML]{EFEFEF}\textbf{0.49} & \cellcolor[HTML]{EFEFEF}\textbf{0.55} & \cellcolor[HTML]{EFEFEF}\textbf{0.54} & \cellcolor[HTML]{EFEFEF}\textbf{0.50} & \cellcolor[HTML]{EFEFEF}\textbf{0.52} & \cellcolor[HTML]{EFEFEF}\textbf{0.49} & \cellcolor[HTML]{EFEFEF}\textbf{0.53} & \cellcolor[HTML]{EFEFEF}\textbf{0.50} & \cellcolor[HTML]{EFEFEF}\textbf{0.56} & \cellcolor[HTML]{EFEFEF}\textbf{0.55} & \cellcolor[HTML]{EFEFEF}\textbf{0.53} & \cellcolor[HTML]{EFEFEF}\textbf{0.56} \\
\multicolumn{1}{l}{\multirow{-2}{*}{GitHub}} & \multicolumn{1}{l}{Random} & 0.47 & 0.53 & 0.53 & 0.49 & 0.49 & 0.48 & 0.51 & 0.42 & 0.52 & 0.53 & 0.51 & 0.53 \\ \hline
\end{tabular}
}
\end{table*}

\section{Discussion}
In this Section we will discuss how the proposed algorithm improves different GNN models for both node and link classification problems.

\paragraph{Node Classification: } GCN \cite{kipf2017semisupervised} is one of the pioneering works on GNN, which uses a convolution-based first-order neighbourhood aggregation. GCN with our generated features produced highest accuracy for 9 out 10 data sets while for computer network dataset the result is less than that of random. The improvement ranges from 2.3\% to 12.6\%. The SAGE model proposed in \cite{10.5555/3294771.3294869} moves away from the transductive setting of GNN to generate node embeddings for unseen nodes using inductive neighbourhood aggregation. We can see, from Table \ref{tab:table2}, that the proposed feature generates best accuracy for node classification problem for 9 out of 10 data sets with a mean increase of $2.99\%$ when the SAGE model is initialized with our proposed feature set. Interestingly, for CiteSeer data set the improvement for node classification problem is over 21\%. These works were followed a year later by GAT \cite{veličković2018graph}, where the authors introduced attention to the neighbourhood features before aggregation. We obtain a mean increase of $3.42\%$ and highest improvement of $11.39\%$. In ResGated \cite{bresson2018residual}, the authors use the edge gating mechanism along with the residual embedding of a node. They extended TreeLSTM \cite{tai-etal-2015-improved} in the pipeline while solving the limitations of it. We obtain improved results for 8 out of 10 data sets with improvement ranging from 1.4\% in ES and 27.27\% in CiteSeer. All the models introduced so far look at GNNs as a back box lacking a critical understanding of their working and the areas in which they can fail.
The authors of GraphConv \cite{10.1609/aaai.v33i01.33014602} look at GNNs from a theoretical point of view and try to draw a relation between GNNs and the WL isomorphism test. The authors propose K-dimensional GNN, which are more powerful than graph-based neural networks. Similar to other methods here also highest improvement found in CiteSeer data with improvement of 30.66\%. However, our method produce low accuracy on Flicker data. Overall it produced best results for eight out of ten data sets.
Authors of all the papers mentioned above have manually designed their architecture given the task at hand. The paper General \cite{10.5555/3495724.3497151} proposes a novel method for designing GNNs for different tasks by studying the different architectural choices in designing GNNs. This reduces the dependency of the existing models on the manual architecture search. We use a model from the proposed design space and achieve a mean increase of $3.21\%$ with best results for seven data sets. In traditional message-passing graph networks, the representation of the source node is considered during message propagation while ignoring the target node. In FiLM \cite{10.5555/3524938.3525045}, the authors consider the representation of the target node during message propagation. Here also a mean improvement of 5.11\% is observed when we use the proposed features in the mode. In SSP \cite{izadi2020optimization}, the authors use natural gradients to optimize GNNs. The authors of SSP hold a record for the highest node classification accuracy on Cora using the old features. We have improved their existing results and have set the new state of the art for the Cora dataset at the accuracy of $0.92$. We only report the results of SSP for node classification on the data sets shown in their paper.

Graph Transformers \cite{ijcai2021p214} are more recent developments in GNN inspired by the original Transformers and encode certain positional and structural information into the graph encoding. These encodings make the gene structure and position-aware, increasing its representation capability. Structural and positional awareness helps in the attention mechanism, which is challenging to scale for large graphs. We obtain a mean increase of $3.2\%$. GAT discussed earlier didn't depend on the query note formally defined as static attention in GAT2 \cite{brody2022how}. The authors of GAT2 introduce dynamic attention by conditioning attention on the query node, which increases the expressiveness of the final embedding. We see over $4\%$ mean increase here. The recent most model Gen \cite{10225480} introduce a generalised and differentiable aggregation function that is also permutation invariant. Unlike normal aggregation functions like mean max or average generalised aggregation, Gen has learnable parameters that are trained task-specific along with the GNN. We obtain a mean accuracy increase of $9.24\%$ for the node classification problem.

The proposed features have improved all of the models across different data sets. One of the fact we identified that for PubMed data the best accuracy obtained by the actual encoding for 10 out 12 models. Only two models able to improve over the actual encoding are GATv2 and TranSage. Interestingly, the best results obtained here when the actual features are augmented with the proposed generated features. On the other hand, the improvement observed in CiteSeer data set is very high ranging from $11\%$ to $44\%$ across different models.

\paragraph{Link Classification:} Similar to the node classification problem, our proposed features generated improved results for link classification problem. The observation from Table \ref{tab:table3} reveal that for GitHub, Computer, PT, FR and CiteSeer, the proposed generated features improved the accuracy for majority of the model. While for PubMed data the produced features either beat the other methods or generated joint best accuracy for all the models except FiLM model. In case of Cora and ES we got mixed results. However, for Cora data overall highest accuracy is obtained by the model General while using the generated features.

\section{Ablation Study}
\label{sec:ablation}
\subsection{Dynamic Network}
We know that GNNs, such as GCNs, can handle incoming nodes. In this Section, we show the tolerance of our proposed feature generation model to new nodes in a dynamic network. This makes the image model inductive as it is unaware of the neighbouring nodes. We take the Cora dataset as the basis for this study. We split the Cora graph nodewise into $5$ subgraphs $S_1$, $S_2$, $S_3$, $S_4$, $S_5$ with a split size of $50\%$, $10\%$, $10\%$, $10\%$, $10\%$, $10\%$ respectively. In the first step, we train our image generation model on multiple energy plots for each node of subgraph $S_1$, extract the features from the model, finetune extracted features using a inductive GNN and test on the subgraph $S_2$. In the next step, we extract plots from subgraphs of nodes in $S_2$ and finetune the trained image model on these plots, extract features of all nodes present in subgraphs $S_1$ and $S_2$ and then test using the same procedure on subgraph $S_3$. This is repeated till we finetune on subgraph $S_4$ and test on $S_5$. The results are shown in Table \ref{tab:table4}. As expected the accuracy is decrease with more and more new nodes are added to the network. In the case of edge deletion, the image model needs to be fed with new neighbourhood images of the nodes the deleted edges affect. 
\begin{table}[h]
\centering
\caption{The performance of generated features using the models in a Dynamic Graph Scenario. T@k denotes Training on k\% nodes. The results presented are for testing done on 10\% of unseen nodes.}
\label{tab:table4}
\resizebox{0.7\columnwidth}{!}{%
\begin{tabular}{llllll}
\hline
 & \textbf{T@50} & \textbf{T@60} & \textbf{T@70} & \textbf{T@80} & \textbf{T@90} \\\hline
M1 & 0.51 & 0.34 & 0.30 & 0.30 & 0.33 \\
M2 & 0.75 & 0.60 & 0.51 & 0.43 & 0.44 \\
M3 & 0.75 & 0.61 & 0.47 & 0.43 & 0.42 \\
M4 & 0.47 & 0.39 & 0.30 & 0.30 & 0.33 \\
M5 & 0.77 & 0.67 & 0.58 & 0.50 & 0.45 \\
M6 & 0.50 & 0.33 & 0.30 & 0.30 & 0.33 \\
M7 & 0.76 & 0.61 & 0.52 & 0.43 & 0.44 \\
M8 & 0.77 & 0.75 & 0.79 & 0.67 & 0.61 \\
M9 & 0.75 & 0.56 & 0.42 & 0.39 & 0.36 \\
M10 & 0.77 & 0.73 & 0.74 & 0.65 & 0.59 \\
M11 & 0.76 & 0.61 & 0.51 & 0.43 & 0.44 \\
\hline
\end{tabular}%
}
\end{table}
\subsection{Changing the neighbourhood size for a node}
\begin{figure}[t]
    \centering
    \includegraphics[width=\columnwidth]{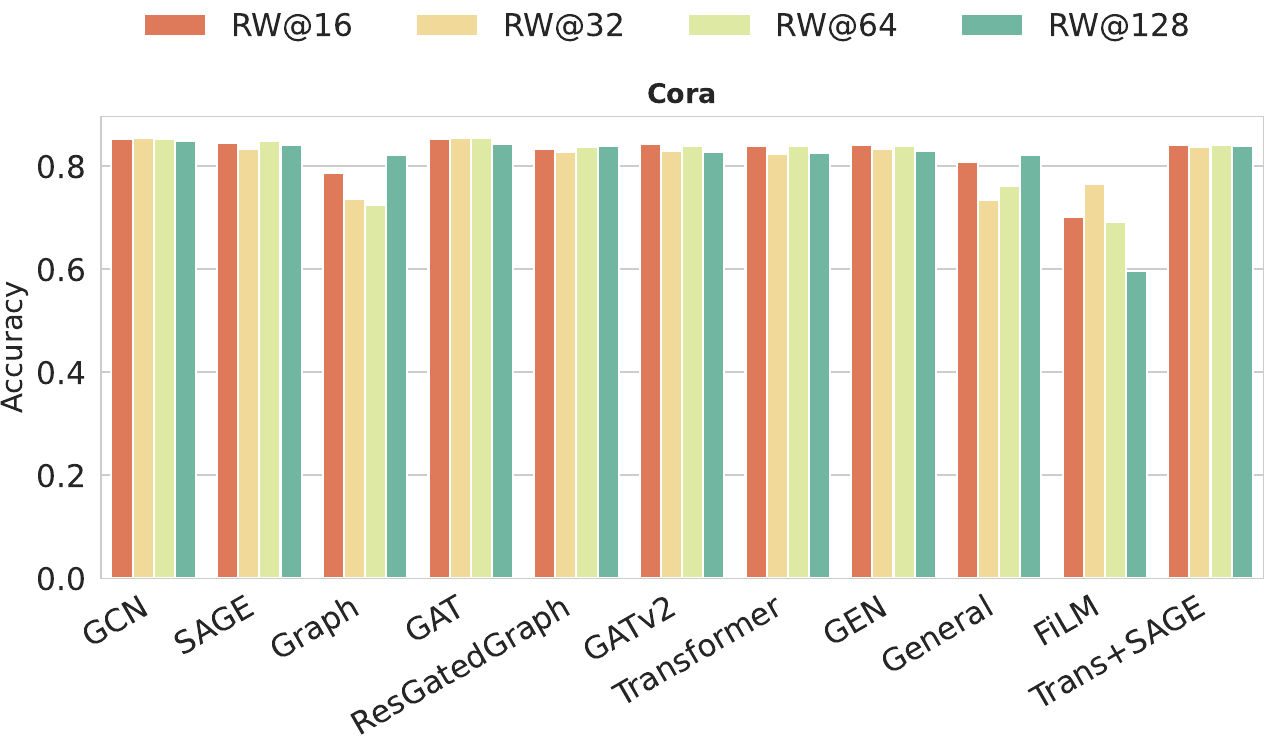} 
    \caption{Performance of the Models on the Cora Dataset on changing the size of the Random Walk. Here, RW@16 denotes a random walk of size 16. }
    \label{fig:fig3}
\end{figure}

In this Section, we try to identify the change in classification accuracy across the models when we change the local neighbourhood size of a node. We can control the local neighbourhood size of a node by setting an upper bound on the length of the random walk for that node. We show the node classification accuracy on the Cora dataset for different neighbourhood sizes in Figure \ref{fig:fig3}. 
\subsection{Tolerance to training size}
In most of the results we presented earlier, we trained our image model and all GNNs transductively on $80\%$ of the nodes and tested on $20\%$ nodes as it is a popular split ratio in deep learning literature. In this section, we check the tolerance of our method to a decreasing training size. We take $80:20$, $70:30$, $60:40$ and $50:50$ as our train test split ratios. This study shows the scalability of our generation model in situations where we train on half the dataset consisting of the essential nodes and test on the remaining. As usual, we use the Cora dataset for the study. The results of this study are presented in Table \ref{tab:table5}. We can see from the results across the models that we gain a maximum of $8\%$ accuracy on training on $50\%$ to training on $80\%$ of the nodes. This shows that our method is quite robust to a decreasing number of training nodes.
\begin{table}[h]
\caption{Comparison of Models Trained on Different Train and Test Splits of the Cora dataset. Here S1 represents `Generated Features', and S2 represents `Generated+Actual features'}
\label{tab:table5}
\resizebox{\columnwidth}{!}{%
\begin{tabular}{llrrrrrrrrrrr}
\hline
 &  & \multicolumn{1}{l}{M1} & \multicolumn{1}{l}{M2} & \multicolumn{1}{l}{M3} & \multicolumn{1}{l}{M4} & \multicolumn{1}{l}{M5} & \multicolumn{1}{l}{M6} & \multicolumn{1}{l}{M7} & \multicolumn{1}{l}{M8} & \multicolumn{1}{l}{M9} & \multicolumn{1}{l}{M10} & \multicolumn{1}{l}{M11} \\ \hline
\multirow{2}{*}{\begin{tabular}[c]{@{}l@{}}Split 1\\ (50, 50)\end{tabular}} & S1 & 0.83 & 0.81 & 0.83 & 0.82 & 0.83 & 0.83 & 0.82 & 0.78 & 0.83 & 0.80 & 0.82 \\
 & S2 & 0.83 & 0.86 & 0.86 & 0.86 & 0.86 & 0.85 & 0.85 & 0.79 & 0.86 & 0.83 & 0.86 \\ \hline
\multirow{2}{*}{\begin{tabular}[c]{@{}l@{}}Split 2\\ (60,40)\end{tabular}} & S1 & 0.84 & 0.83 & 0.84 & 0.84 & 0.83 & 0.84 & 0.83 & 0.80 & 0.85 & 0.82 & 0.83 \\
 & S2 & 0.87 & 0.86 & 0.86 & 0.85 & 0.86 & 0.84 & 0.86 & 0.80 & 0.85 & 0.86 & 0.86 \\ \hline
\multirow{2}{*}{\begin{tabular}[c]{@{}l@{}}Split 3\\ (70,30)\end{tabular}} & S1 & 0.87 & 0.86 & 0.87 & 0.85 & 0.87 & 0.86 & 0.86 & 0.84 & 0.86 & 0.84 & 0.86 \\
 & S2 & 0.88 & 0.87 & 0.85 & 0.82 & 0.87 & 0.84 & 0.87 & 0.82 & 0.86 & 0.86 & 0.87 \\ \hline
\multirow{2}{*}{\begin{tabular}[c]{@{}l@{}}Split 4\\ (80,20)\end{tabular}} & S1 & 0.86 & 0.86 & 0.87 & 0.86 & 0.87 & 0.86 & 0.87 & 0.85 & 0.87 & 0.84 & 0.86 \\
 & S2 & 0.89 & 0.89 & 0.88 & 0.89 & 0.88 & 0.89 & 0.89 & 0.84 & 0.88 & 0.88 & 0.90 \\ \hline
\end{tabular}%
}
\end{table}

\subsection{Why DenseNet?}
We have used DenseNet to extract the node features from the images. We have also experimented with other popular image models like  Resnet18, Resnet152, Alexnet, VGG11, VGG16, Squeezenet, Inception and vision transformers (viT). We obtain the best training results using DenseNet. The training accuracy across all the datasets is more than $90\%$, and the test accuracy varies depending on the number of classes for prediction from $40\%$ for $8$ class classification to $67\%$ for two class classifications. 

\section{Related Work}
Early works on embedding graph neural networks used shallow embedding methods that used factorization, like Word2Vec \cite{Mikolov2013EfficientEO} and Matrix Factorization \cite{koren2009matrix}, to adapt to sparse data. These factorization methods inspired the researchers of DeepWalk \cite{10.1145/2623330.2623732}, and Node2Vec \cite{10.1145/2939672.2939754} to factorize graph nodes with node embedding vectors. The random walk-based method considers a multi-order neighbourhood structure of the nodes of a graph. Authors of \cite{10.1145/2736277.2741093} use other objective functions to encode the first and second-order graph neighbourhood structures efficiently for various graphs. Once we get the embeddings for a node, it becomes essential to scale the methods, such as Node2Vec, to larger graphs as done in \cite{lerer2019pytorch} and \cite{10.1145/3308558.3313508}. Graph neural networks follow the Node2Vec implementations. A GNN works on the principle of message propagation and message aggregation. A node is passed the information of its neighbours, and it aggregates them along with its information to produce a new embedding. The success of GNNs over traditional Node2Vec methods led to successive works in this field, like GCN \cite{kipf2017semisupervised}, which redesigned the popular image convolution method into graphs, and GATs \cite{veličković2018graph} which give a preferential aggregation of information from the neighbours using an attention score for each neighbour. Recent developments in GNNs are GraphTransformers \cite{NEURIPS2019_9d63484a}, which, inspired by the original transformers, encode certain positional and structural embeddings in the graph encoding to make the GNN's position and structure-aware. This structural and positional awareness helps the attention mechanism but is challenging to scale for large graphs. At this point, the work on GNNs is quite diversified in multiple broader directions like node classification, link prediction and community detection. Researchers in all these fields try to simplify the GCN structure by removing feature transformations and non-linear activations in papers like \cite{10.1145/3397271.3401063} and \cite{10.1145/3543507.3583340}. Some common problem plagues all the existing GNN methods across the tasks, like GNNs cannot count triangles or distinguish autormorphic graphs \cite{NEURIPS2021_7ffb4e0e}. GNNs can be considered as powerful as Weisfeiler and Leman test but can still assign different embeddings to isomorphic graphs \cite{NEURIPS2019_bb04af0f, xu2018how}. It is challenging to scale GNNs as, in most cases, the embeddings of each node are also a training parameter, and there can be many neighbours of a node. A large number of neighbours is commonly seen in the case of social network graphs. In this paper, we try to address the problems of encoding the structural information into an embedding of a node in an implicit batched manner. This technique also does not require training the embedding. It only updates the network weights using a GNN, as the embeddings already contain sufficient structural information for node classification and link prediction tasks. 

The literature for visualizing graphs mainly uses a force-directed method \cite{6295853, 7557906, 6658169, 6902873} for graph drawing or is wholly based on the Kamadi-Kawai \cite{8285432} algorithm used in our paper. Some recent works in this field use GNNs \cite{9810169} for graph visualization to enhance representations and minimize edge crossing in graphs. We have selected the Kamadi-Kawai \cite{kamada1989algorithm} algorithm as it produced consistent results across the datasets with minimum time complexity.

\section{Conclusion}
This study described a novel methodology GraphViz2Vec to generate structure-aware feature of nodes that can be used in downstream tasks with GNN models. Further, we showed, once the feature is generated only 2 layers of GNN is sufficient to produce good results for node and link classification problem. This solve the problem of over-smoothing. The claims are supported with extensive experiments on 10 data sets with 12 different GNN models. The work is the first use of minimum energy based graph visualization to generate node level features to best of our knowledge. This approach opens a new avenue that can provide direction to scale GNN for larger graphs by leveraging the abilities of deep learning vision models. 

\bibliographystyle{ACM-Reference-Format}
\bibliography{sample-base}

\appendix
\section{Appendix}
\subsection{Changing the size of hidden state}
We have experimented with different hidden state sizes for the results presented in Table \ref{tab:table2} and Table \ref{tab:table3}. In this Section, we show the change in the model accuracy and loss for different hidden states on the Cora dataset. The results are shown in Figure \ref{fig:fig1}. We can see that the loss converges after $40$ epochs. The loss is higher for a bigger hidden state in most models. We can see that a hidden state of $256$ is sufficient for most cases.

\begin{figure}[b]
    \centering
    \includegraphics[width=0.6\columnwidth]{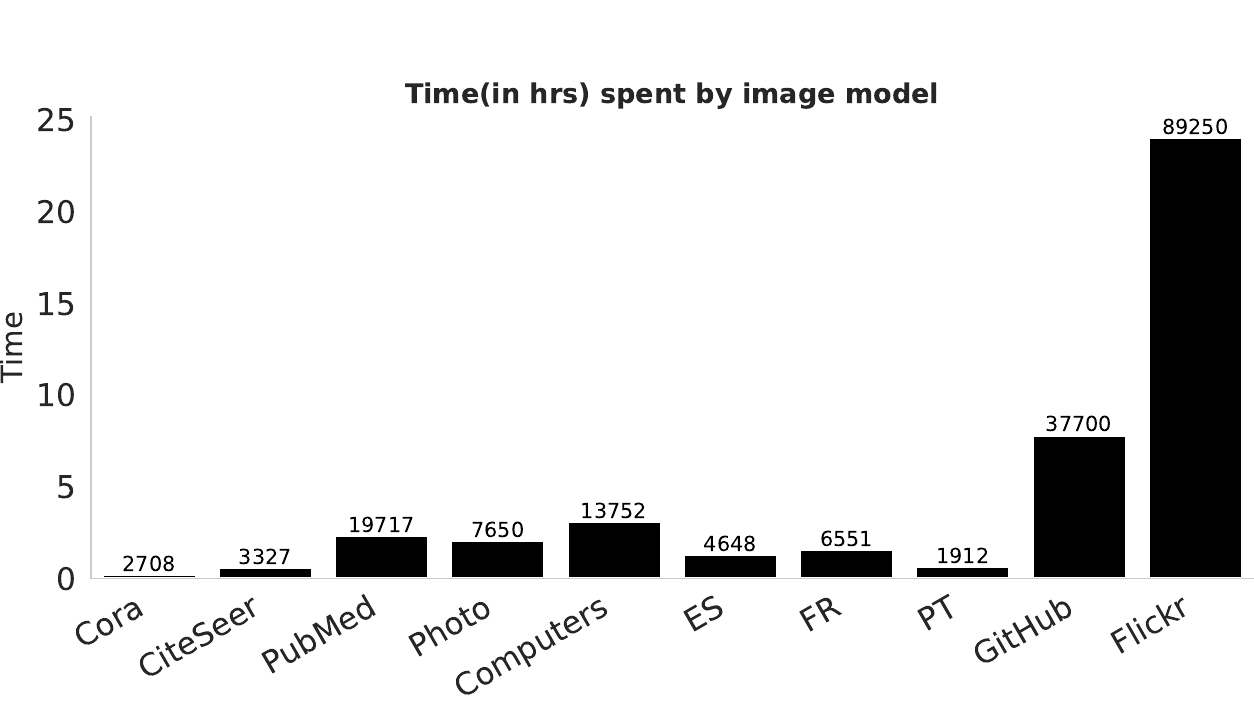} 
    \caption{Time(in hrs) spent on training by image model for each dataset. We have taken one image per node in this case. The labels on each bar denote the number of nodes in that dataset.}
    \label{fig:fig4}
\end{figure}

\subsection{Time spent on training}
We have shown the time spent training the GNNs in Figure \ref{fig:fig2} for all the datasets across all the models. We also show the time spent training the image model (DenseNet) for all the datasets in Figure \ref{fig:fig4}. The training set for the image model contains a single image per node for all the datasets. The numbers on each bar represent the number of nodes in that dataset. The image model is found to converge on $70$ epochs for training across all the datasets.

\begin{figure}[b]
    \begin{subfigure}{\linewidth}
        \centering
        \includegraphics[width=\linewidth]{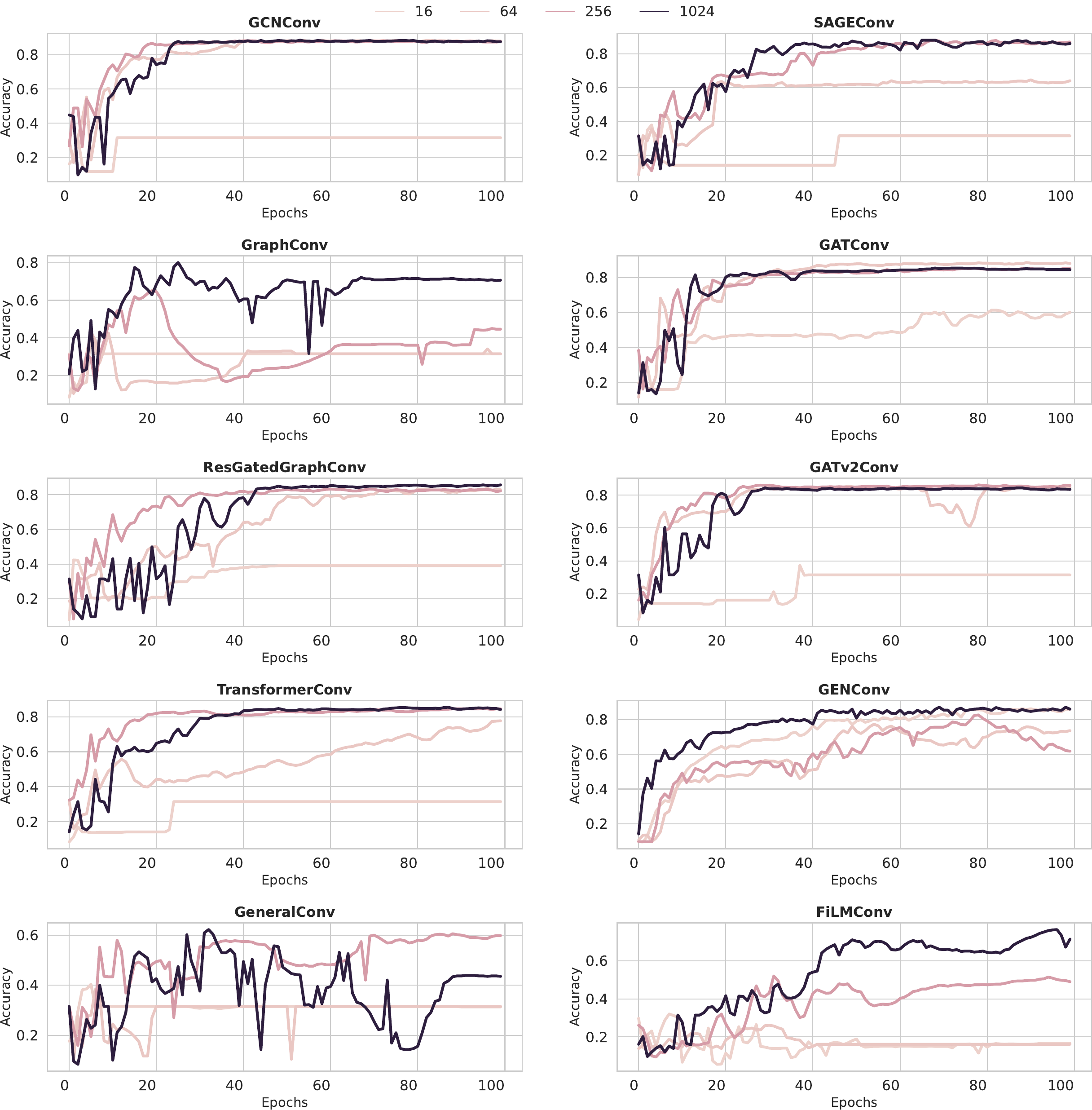}
        \caption{Accuracy}
        \label{fig:image_1a}
    \end{subfigure}
    \hfill
    \begin{subfigure}{\linewidth}
        \centering
        \includegraphics[width=\linewidth]{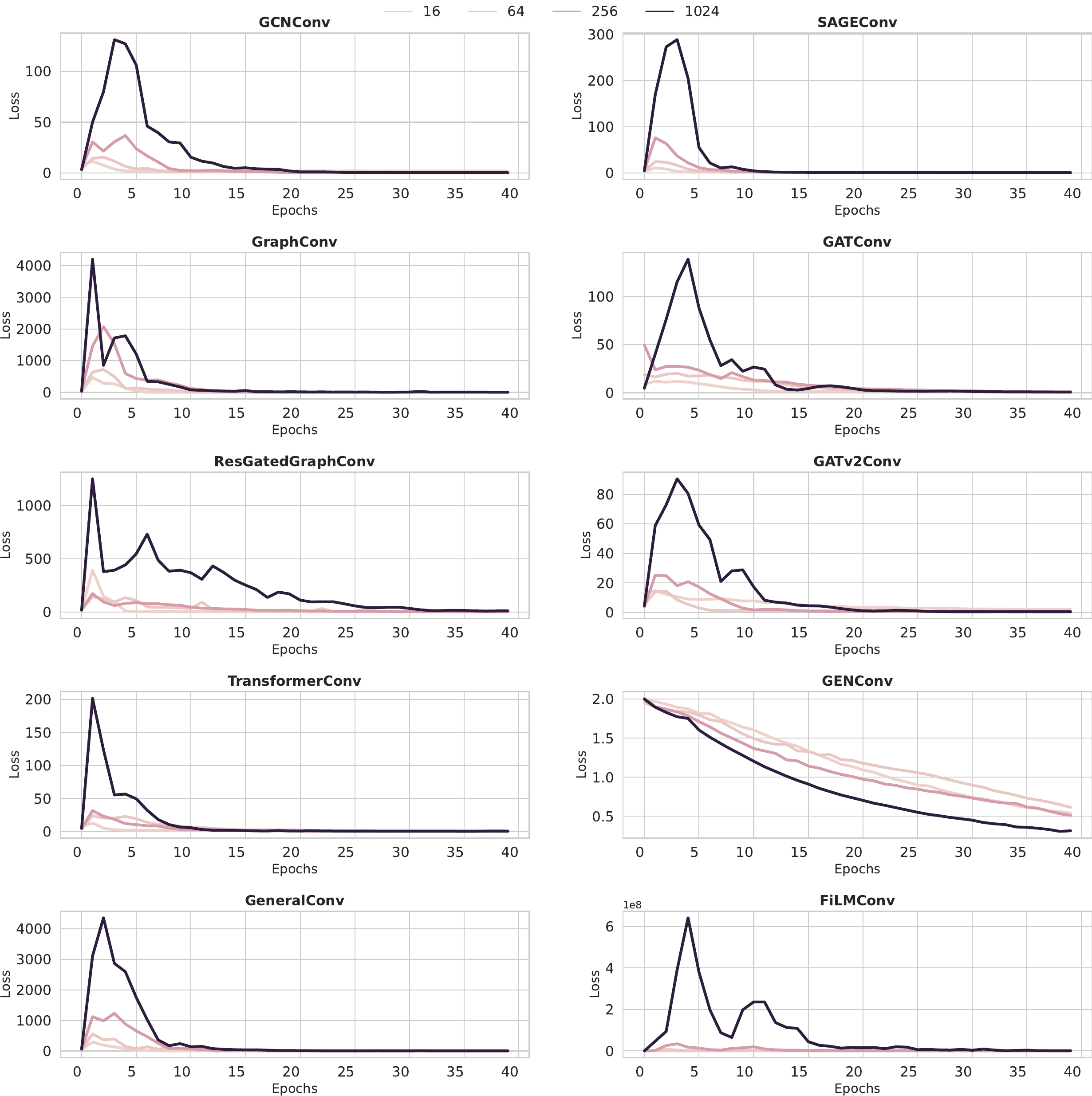}
        \caption{Loss}
        \label{fig:image_1b}
    \end{subfigure}
    \caption{Change in Accuracy(a) and Loss(b) through the increase in epochs with different hidden feature sizes for all the $2$ layer models on the Cora Dataset (the images are generated from a random walk of size 128).}
    \label{fig:fig1}
\end{figure}

\begin{figure*}[]
    \centering
    \includegraphics[width=\linewidth]{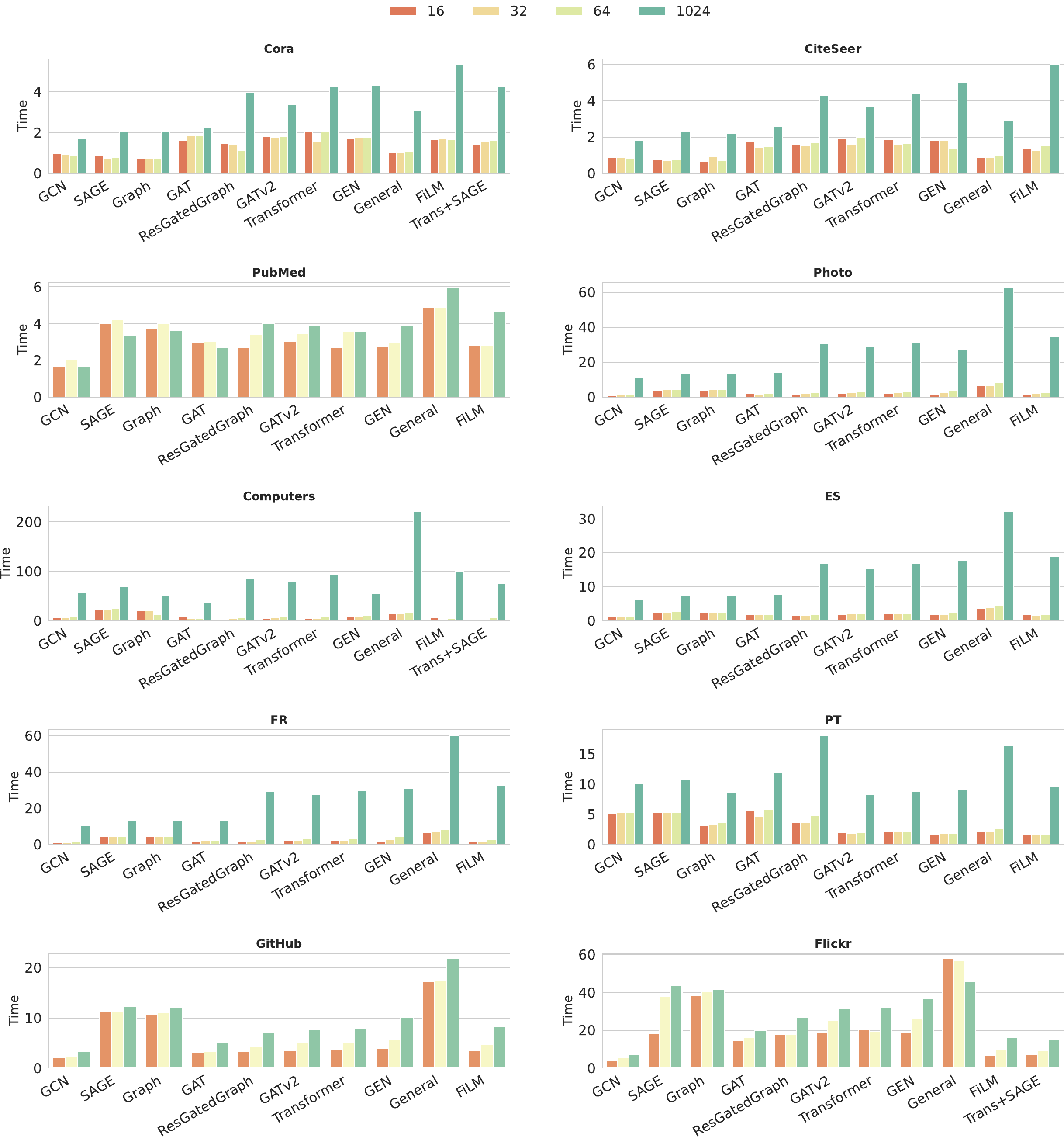} 
    \caption{Time(in seconds) taken by the different models for different hidden states when trained on the generated features to reach 600 epochs}
    \label{fig:fig2}
\end{figure*}

\end{document}